\documentclass[preprint,12pt]{elsarticle}

\usepackage{amssymb}
\usepackage{amsmath}
\usepackage{multirow}
\usepackage{threeparttable}
\usepackage{hyperref}
\hypersetup{hidelinks,
	colorlinks=true,
	allcolors=black,
	pdfstartview=Fit,
	breaklinks=true}

\journal{Automation in Construction}

\begin{document}

\begin{frontmatter}


\title{Topology-aware Mamba for Crack Segmentation in Structures\tnoteref{label1}}
\author[1]{Xin Zuo}
\author[1]{Yu Sheng}
\author[2]{Jifeng Shen\corref{wcor1}}
\author[3]{Yongwei Shan\corref{wcor1}}

\cortext[wcor1]{Corresponding Author}


\affiliation[1]{organization={School of Computer Science and Engineering},
             addressline={Jiangsu University of Science and Technology},
             city={Zhenjiang},
             postcode={212003},
             country={China}}
\affiliation[2]{organization={School of Electronic and Informatics Engineering},
	addressline={Jiangsu University},
	city={Zhenjiang},
	postcode={212013},
	country={China}}   
\affiliation[3]{organization={School of Civil and Environmental Engineering},
	addressline={Oklahoma State University},
	city={Stillwater, OK},
	postcode={74074},
	country={USA}}  




\begin{abstract}

CrackMamba, a Mamba-based model, is designed for efficient and accurate crack segmentation for monitoring the structural health of infrastructure. Traditional Convolutional Neural Network (CNN) models struggle with limited receptive fields, and while Vision Transformers (ViT) improve segmentation accuracy, they are computationally intensive. CrackMamba addresses these challenges by utilizing the VMambaV2 with pre-trained ImageNet-1k weights as the encoder and a newly designed decoder for better performance. To handle the random and complex nature of crack development, a Snake Scan module is proposed to reshape crack feature sequences, enhancing feature extraction. Additionally, the three-branch Snake Conv VSS (SCVSS) block is proposed to target cracks more effectively. Experiments show that CrackMamba achieves state-of-the-art (SOTA) performance on the CrackSeg9k and SewerCrack datasets, and demonstrates competitive performance on the retinal vessel segmentation dataset CHASE\underline{~}DB1, highlighting its generalization capability. The code is publicly available at: \href{https://github.com/shengyu27/CrackMamba.}{https://github.com/shengyu27/CrackMamba.} 

\end{abstract}




\end{frontmatter}



\section{Introduction}
Cracks are common structural issues that pose a potential threat to the safety and integrity of facilities. Structural deformation, environmental hazards, and various other factors may accelerate crack development, leading to serious safety consequences \cite{1}. Consequently, the implementation of regular, meticulous inspections is paramount for the timely detection and remediation of structural damage. Structural health monitoring (SHM) stands out as one of the primary methods for non-destructive assessment of infrastructure \cite{2}. In this context, experts analyze the state of cracks to determine appropriate solutions for different cracking situations \cite{3}\cite{4}. To expedite this process and alleviate the workload of experts, efficient and accurate segmentation of cracks is imperative.

Initial approaches to crack detection primarily employed threshold segmentation \cite{5}, edge detection \cite{6}, and morphological operations \cite{7}. These traditional methods, however, exhibited limitations in addressing complex image backgrounds, noise interference, illumination inconsistency, and optimal threshold determination \cite{8}. Consequently, they frequently failed to achieve satisfactory crack segmentation outcomes, resulting in suboptimal performance efficacy. Recent advancements in deep learning and computer vision have facilitated the application of image segmentation models to defect detection. Crack propagation, characterized by shear stress-induced expansion and curvature, often results in serpentine trajectories \cite{9}. Accurate and comprehensive crack segmentation for subsequent remediation necessitates the capture of global image features. However, the inherently limited receptive field of CNNs has constrained their performance in crack segmentation tasks. 

Various approaches have been proposed to address this limitation. CrackSegNet \cite{10} employs dilated convolution and spatial pyramid pooling for multi-scale feature fusion to expand the model’s receptive field. DeepCrack \cite{11} proposes a pyramidal feature aggregation network and employs conditional random field post-processing to further refine segmentation results. CrackNet-V \cite{12} employs the stacking of multiple 3 × 3 and 15 × 15 convolutional layers for deep abstraction.  Even with advancements like DeepLabV3's \cite{13} adoption of the Atrous Spatial Pyramid Pooling(ASPP) multi-scale feature fusion structure, this issue remains unresolved. Subsequently, the self-attention mechanism of ViT \cite{14} enables the model to capture global feature information, extending its receptive field to cover the entire image. DTrC-Net \cite{15} employs a dual-branching encoder consisting of both Transformer and CNN, with feature fusion in the decoder. This approach achieves the modeling of global contextual information while accurately extracting low-level feature details. PCTNet \cite{16} proposes a dual-attention network combining local multi-attention and global pooled multi-attention, aiming to reduce the computational cost of Transformer while maintaining competitive segmentation performance. However, the high computational demands of ViT pose a significant obstacle to the real-world application of ViT-based models.

Recently, structured state space sequence models(SSMs) have showcased the potential for efficiency and effectiveness in modeling long-range dependencies, and the proposal of Mamba \cite{17} has made the application to the image domain a possible alternative foundational model to replace Transformers. Benefiting from the low computational costs of Mamba's linear growth according to sequence length, as opposed to the sub-square growth of Transformers, Mamba-based image processing models are gradually demonstrating their potential dominance in the image domain. They may even emerge as the new predominant vision based model in the future \cite{18}\cite{19}. Particularly in the realm of medical image segmentation, numerous studies have emerged to highlight its promise \cite{20}\cite{21}\cite{22}\cite{23}. For instance, Swin-UMamba \cite{22} proposed a network model approximating the Swin-transformer structure and employing VMamba as the backbone and achieved competitive performance with pre-trained weights. Mamba-UNet \cite{23} also explored a Mamba block model with a structure akin to UNet for medical image segmentation. However, thus far, no Mamba-based application model for crack image segmentation has emerged.

Mamba was initially applied to one-dimensional sequences in the field of natural language processing (NLP). While VMamba has been successfully extended to images by patching 2D images into sequences, it still faces certain challenges. The unidirectional sorting of sequences imposes limitations on the receptive field, as highlighted by the 'direction-sensitive' issue raised in VMamba \cite{19}. To address this, VMamba ensures the global receptive field of the model by reordering rows and columns and employing sequences in all four directions for learning before superimposing them. However, VMamba fails to consider the impact of token order within the proximity interval on feature extraction post-reordering. LocalVMamba \cite{24} proposes dividing the local window first and then re-expanding it to prioritize features within the local window, thereby enhancing the model's ability to capture local spatial relationships. PlainMamba \cite{25} adopts a continuous ordering approach to address the problem of spatial discontinuities causing neighboring tokens to decay to varying degrees, exacerbating semantic discontinuities and potentially degrading performance over time. EfficientVMamba \cite{26} employs dilated sequence ordering to maintain performance while achieving model lightweighting, balancing computational efficiency with effectiveness in sequence modeling tasks. However, the sequential ordering methods proposed above essentially follow the same row-and-column format, which may not align well with the serpentine development characteristics of cracks themselves.

\label{fig1}
\begin{figure}[t]
	\centering
	\includegraphics[scale=0.5]{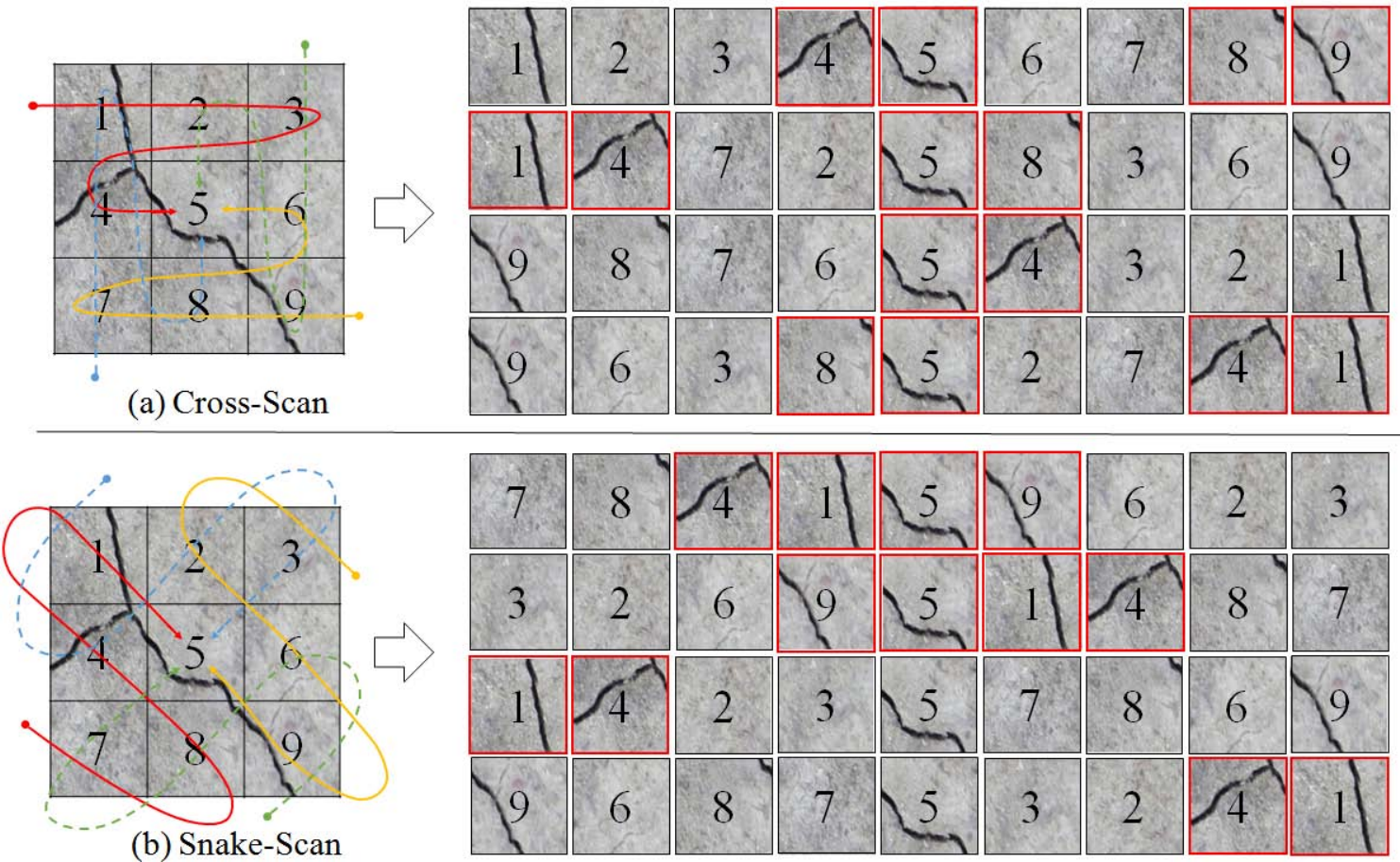}
	
	\caption{(a) is the Cross Scan method, and (b) is our proposed Snake Scan method, with the red border indicating the tokens belonging to semantic proximity.}\label{fig1}
\end{figure}
The main objective of this study is to address the aforementioned challenges in crack segmentation by proposing the CrackMamba model. This proposed model is inspired by Swin-UMamba \cite{22}, the encoder VMambaV2, known for its superior performance in enhancing the model's feature extraction capability. A key innovation in CrackMamba is marked by the introduction of the Snake Scan module, as illustrated in Fig. 1(b), which has been designed differently from the Cross Scan module, Fig. 1(a), used in VMamba. This specialized sequence sorting mechanism has been tailored to the unique characteristics of cracks, enabling denser feature sorting that is more closely aligned with crack semantics. Through this approach, superior feature extraction by Mamba is facilitated, and the model's global modeling capacity is enhanced. Additionally, considering that cracks may exhibit horizontal or vertical trajectories, a new block is designed that combines the original scan method, the Cross Scan by rows and columns, with the new Snake Scan method. This ensures the extraction of both original horizontal and vertical features while enhancing the model's global modeling capability for cracks. Furthermore, to bolster the model's ability to capture local features, a new convolutional branch is also added to the block to integrate local and global feature information. In the experimental section, the results on the different crack datasets demonstrate the effectiveness of our improvements. Moreover, our model exhibits competitive performance in the task of retinal vessel segmentation, which shares similarities with crack segmentation. This further illustrates the excellent generalization capabilities of the model.
To summarize, this paper contributes in the following ways:

\begin{highlights}
	\item A specific sequence scanning method for cracks is proposed, which is superior to the traditional Cross Scan method and enhances the global modeling capability of the model.
	\item A plug-and-play new block for feature extraction that fuses global and local information is proposed.
	\item The first Mamba-based model for crack image segmentation is proposed, which can efficiently and accurately perform crack segmentation and can be generalized to similar segmentation tasks, such as retinal vessel segmentation.
\end{highlights}


\section{Preliminaries}
\subsection{State space models}
Modern SSM-based models, i.e., Structured State Space Sequence Models (S4), all rely on classical continuous systems that map one-dimensional input $x(t) \in \mathbb{R}^{L}$ to sequences $y(t) \in \mathbb{R}^{L}$ via intermediate implicit states $h(t) \in \mathbb{R}^{N}$, which can be expressed as linear ordinary differential equations (ODEs):
\begin{equation}
	\begin{aligned}
		h^{\prime}(t) &= \textbf{A}h(t) + \textbf{B}x(t) \\
		y(t) &=\textbf{C}h(t) 
	\end{aligned} 	
\end{equation}

where $\textbf{A} \in \mathbb{R}^{N \times N}$ denotes the state matrix and $\textbf{B} \in \mathbb{R}^{N \times 1}$ and $\textbf{C} \in \mathbb{R}^{N \times 1}$ denote the projection parameters, which control the dynamic and output mappings, respectively.

The state space model is a continuous time model, and in order to apply it to deep learning algorithms, Eq. (1) is discretized using the zero-order holding technique. The time scale parameter $\Delta$ is introduced and the parameters $\textbf{A}$ and $\textbf{B}$ are converted to discrete parameters $\overline{\textbf{A}}$ and $\overline{\textbf{B}}$ using a fixed discretization rule, defined as follows:
\begin{equation}
	\begin{aligned}
		\overline{\textbf{A}}&=\exp{(\Delta\mathbf{A})} \\
		\overline{\textbf{B}}&=(\Delta\mathbf{\textbf{A}})^{-1}(\exp(\Delta\mathbf{\textbf{A}})-I)\cdot\Delta\mathbf{\textbf{B}}
	\end{aligned}
\end{equation}
After this, Eq. (1) will become:
\begin{equation}
	\begin{aligned}
		h(t)&=\overline{\textbf{A}}h(t-1)+\overline{\textbf{B}}x(t)  \\
		y(t)&=\textbf{C}h(t)
	\end{aligned}
\end{equation}
After discretization, in order to improve the computational efficiency, global operations can be used to parallelize the iterative process described in the accelerated formulation.
\begin{equation}
	\begin{aligned}
		\textbf{y}&=\textbf{x}\odot\overline{\textbf{K}} \\
		\overline{\textbf{K}}&=(\textbf{C}\overline{\textbf{B}},\textbf{C}\overline{\textbf{AB}},...,\textbf{C}\overline{\textbf{A}}^{L-1}\overline{\textbf{B}})
	\end{aligned}
\end{equation}
where $\odot$ represents the convolution operation, $\overline{\textbf{K}}\in\mathbb{R}^L$ and $L$ represents the length of the input sequence.

\subsection{Selective state space models}
Mamba addresses the inherent limitations of the static parameterization in S4 by introducing a Selective State Space Model (S6) that bases its parameters on the input sequences, resulting in a richer, sequence-aware parameterization with improved representational capabilities. Specifically, in the Mamba model, the parameters \textbf{B}, \textbf{C}, and $\Delta$ are derived from the input sequence $x$.

\section{Method}
This section commences with an overview of the model's general structure, as illustrated in Fig. 2. The figure delineates the architecture of the classical segmentation model, which comprises an encoder and a decoder. Following the SwinUMamba setup, VMambaV2 is utilized alongside decoders that are four layers deep. Additional residual connections are incorporated to recover low-level details, and to further enhance upsampling performance, deep supervision is implemented at each upsampling scale. The segmentation head comprises a 2-layer convolutional block responsible for fusing the feature maps, with a 1 × 1 convolutional layer used to output the final segmentation result. Subsequently, concepts related to Mamba are introduced, along with details of each module that has been improved.

\label{fig2}
\begin{figure}[!t]
	\centering
	\includegraphics[scale=0.5]{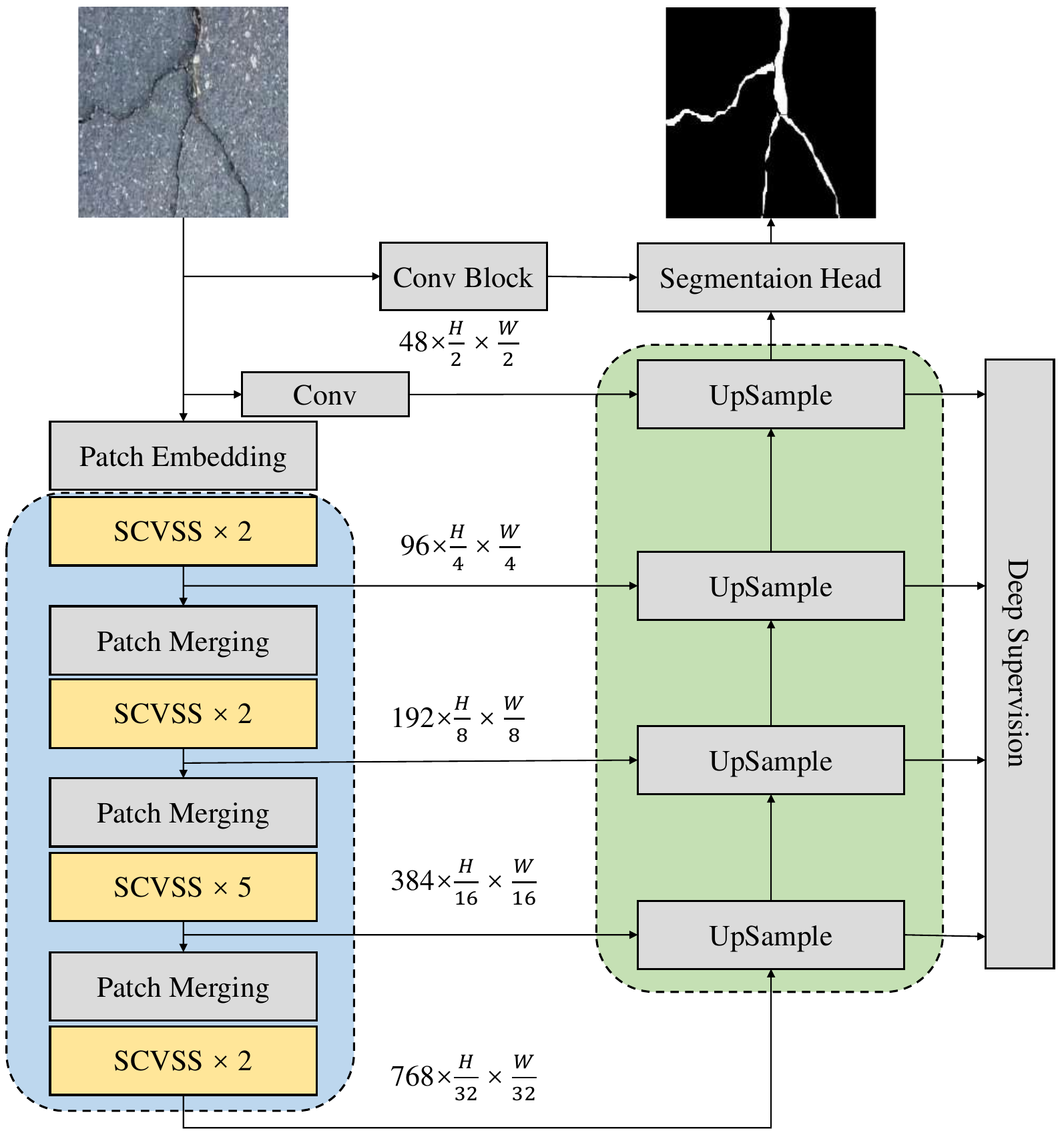}
	\caption{Overall structure of the CrackMamba model.}\label{fig2}
\end{figure}

\subsection{2D-Snake-Scan}
To enable Mamba to better extract crack features, a specialized snake scanning method, which aligns more closely with the characteristics of crack propagation, is designed. Specifically, the snake scanning coordinate indexes of the main diagonal and antidiagonal of images at different scales are first computed to obtain the serpentine sequences in two directions. The image patches are then reordered according to these indexes. The serpentine sequences of the other two directions are subsequently obtained by reversing the direction. This scanning method ensures a global receptive field while better fitting the developmental characteristics of cracks.

The original VSS block takes the feature map $x$, patches it into different sizes, and then generates four different sequences based on the four directions. Each feature sequence is then processed by the SSM, and the final output is merged and reduced to a 2D feature map $\overline{\mathcal{X}_{\mathcal{V}}}$. In contrast, the SnakeVSS block differs from the VSS block in that, when generating the sequences, it creates new sequences different from the latter based on the direction of the snakes. Finally, it reduces to a 2D feature map $\overline{x_{v_s}}$. The above process can be described as:

\begin{align}
		x_{\nu},x_{\nu_{s}}&=expand(x,v,v_{s}) \\
		\overline{x_{v}},\overline{x_{v_{s}}}&=S6(x_{v},x_{v_{snake}}) \\
		\overline{x_{\nu}}&=merge(\overline{x_{1}},\overline{x_{2}},\overline{x_{3}},\overline{x_{4}}) \\
		\overline{{x_{\nu_{s}}}}&=merge(\overline{{x_{s1}}},\overline{{x_{s2}}},\overline{{x_{s3}}},\overline{{x_{s4}}})
\end{align}

Where $v\in V=\{1,2,3,4\}$ represents the four different directions in VMamba, and $v_s\in V_s=\{s1,s2,s3,s4\}$ denotes the four distinct serpentine directions proposed by us. The operations $expand()$ and $merge()$ are responsible for scan expansion and scan merge, respectively, facilitating the division of the image into sequences and the subsequent merging of these sequences back into the image. The S6 serves as the core of the VSS block, enabling each element in the sequence to learn interactively from the compressed hidden state and any previously scanned elements. For further details on S6, please refer to VMamba.

\subsection{Encoder}
Our base module is derived from the VSS block in VMambaV2, upon which an improved version of the SCVSS Block is proposed. This improved version integrates the Snake Scan module SnakeVSS with the base module VSS. Additionally, to further enhance the model's extraction of local feature information, another convolutional branch is added. Spatial attention and channel attention modules are employed to adjust the learning degree of different branches, as depicted in Fig. 3.

\label{fig.3}
\begin{figure}[!t]
	\centering
	\includegraphics[scale=0.5]{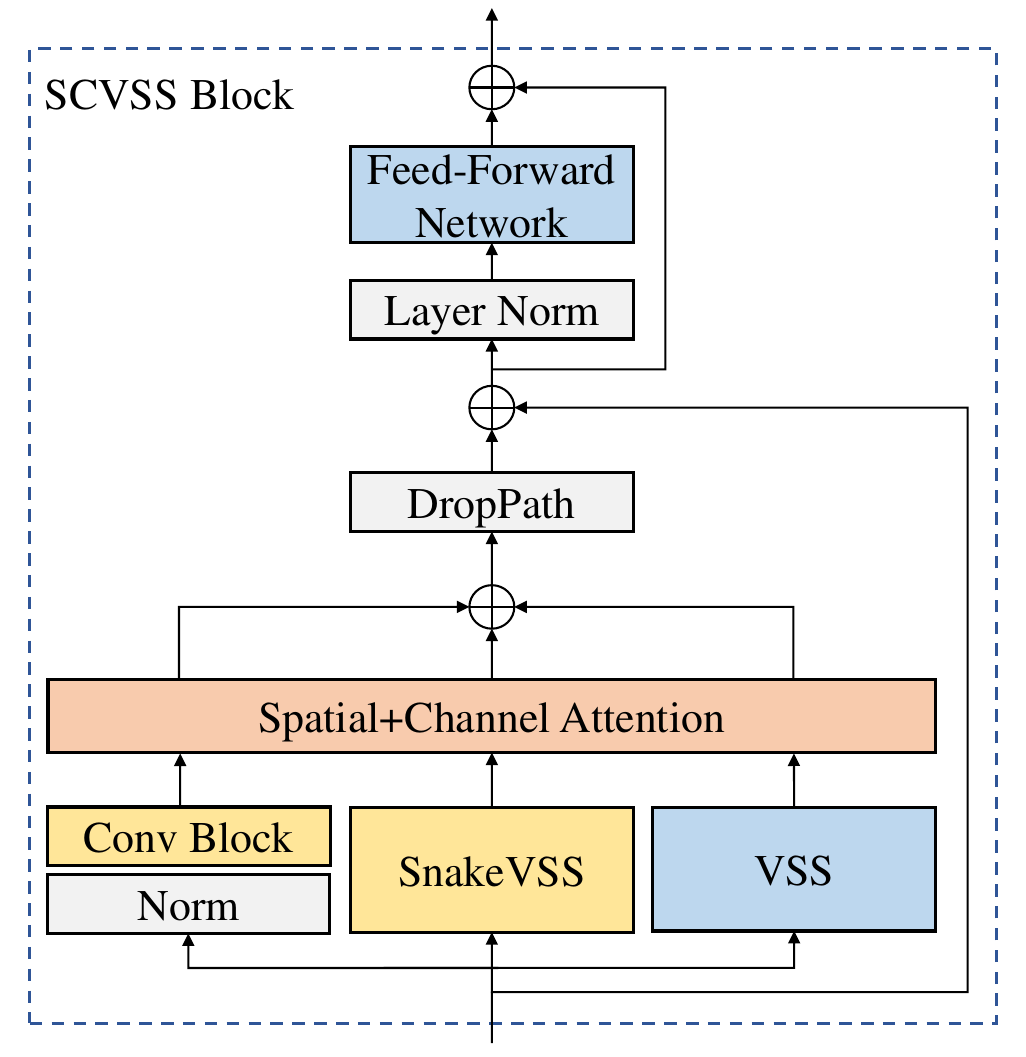}
	\caption{Overall structure of the SCVSS module.}\label{fig.3}
\end{figure}

\subsubsection{SnakeVSS block}

After the feature map enters the SCVSS block, it is divided into three branches: the original VSS block, the SnakeVSS, and the convolution branch. To ensure the extraction of potential horizontal and vertical features of the cracks, the original VSS is retained, utilizing the pre-trained weights from VMambaV2. Simultaneously, considering the developmental characteristics of the cracks, the sequence order of the images is altered using a snake scanning approach before they undergo the Mamba core operation (SS2D), with the specific structure shown in Fig. 4. Additionally, to enhance the model's capability in extracting local feature information, the convolutional branch processes the input feature maps. Specifically, the input feature map $x_{in}$ is processed by the SCVSS block to produce the output $x_{out}$:
\begin{align}
x_{conv}&=CONV(Norm(x))     \\
	x_{snakevss}&=SnakeVSS(x)    \\
	x_{vss}&=VSS(x) \\
x_{out}&=MLP(x+DropPath\big(SCA(x_{conv})
	\notag	\\
	&\quad  +SCA(x_{snkaevss})+SCA(x_{vss})\big))
\end{align}
Where $Norm$ denotes normalization, $CONV$ denotes the convolutional block, $VSS$ refers to the original Mamba core block, $SnakeVSS$ signifies the use of Snake Scan's VSS, $SCA$ represents the Spatial and Channel Attention Module, $DropPath$ indicates the regularization method used to prevent model overfitting, and $MLP$ denotes the forward network.

\label{fig.4}
\begin{figure}[!h]
	\centering
	\includegraphics[scale=0.5]{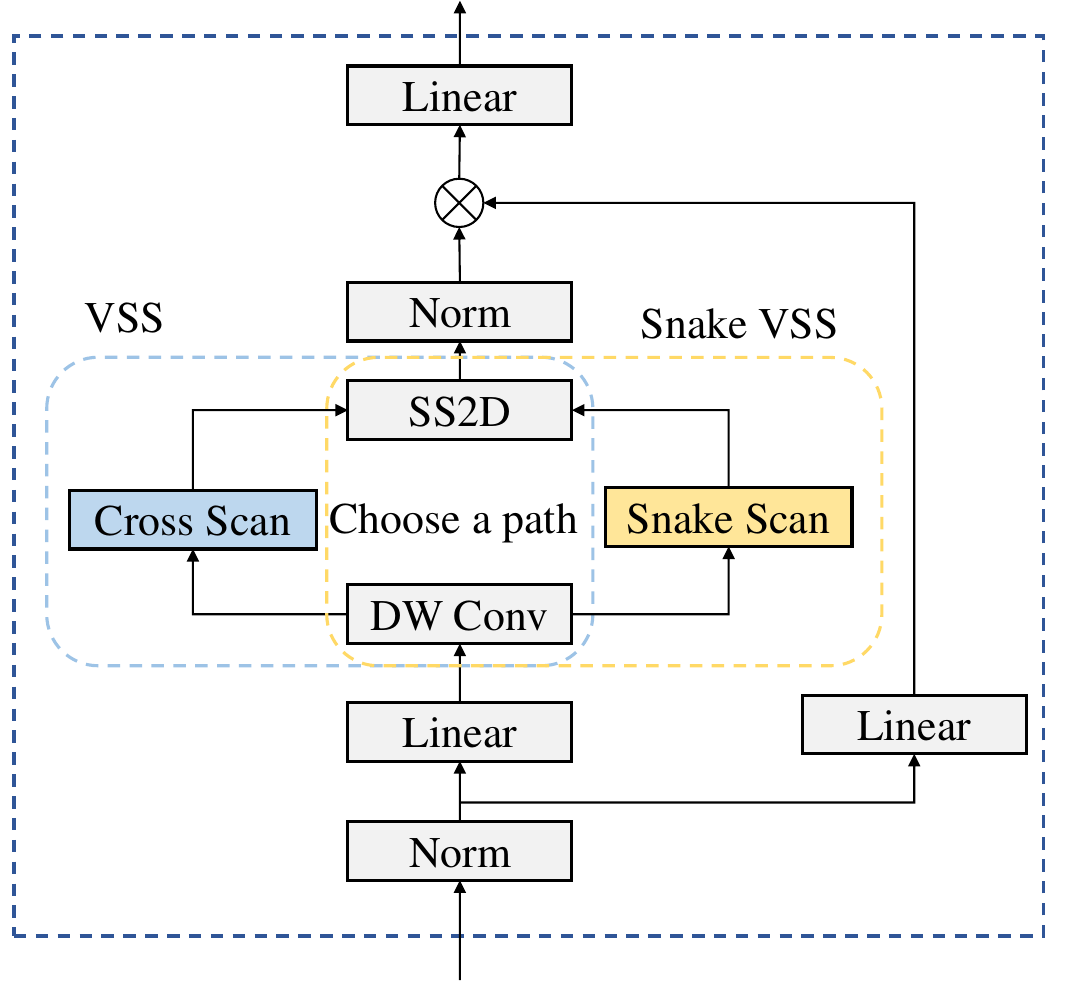}
	\caption{Details of SnakeVSS and VSS structure.}\label{fig.4}
\end{figure}
\label{fig.5}
\subsubsection{Spatial and Channel Attention}
After the feature map is processed through the three branches, it goes through spatial attention to efficiently extract important regions in the map and channel attention to focus on significant features within the channels. Subsequently, they are superimposed. The structural details are shown in Fig. 5. Maximum pooling and average pooling are employed to filter the extraction of spatial regions or channels that require special attention by the model. The final processed feature map $x_{out}$ is obtained from the input feature map $x_{in}$ according to the following equation:
\label{fig.6}
\begin{figure}[!t]
	\centering
	\includegraphics[scale=0.5]{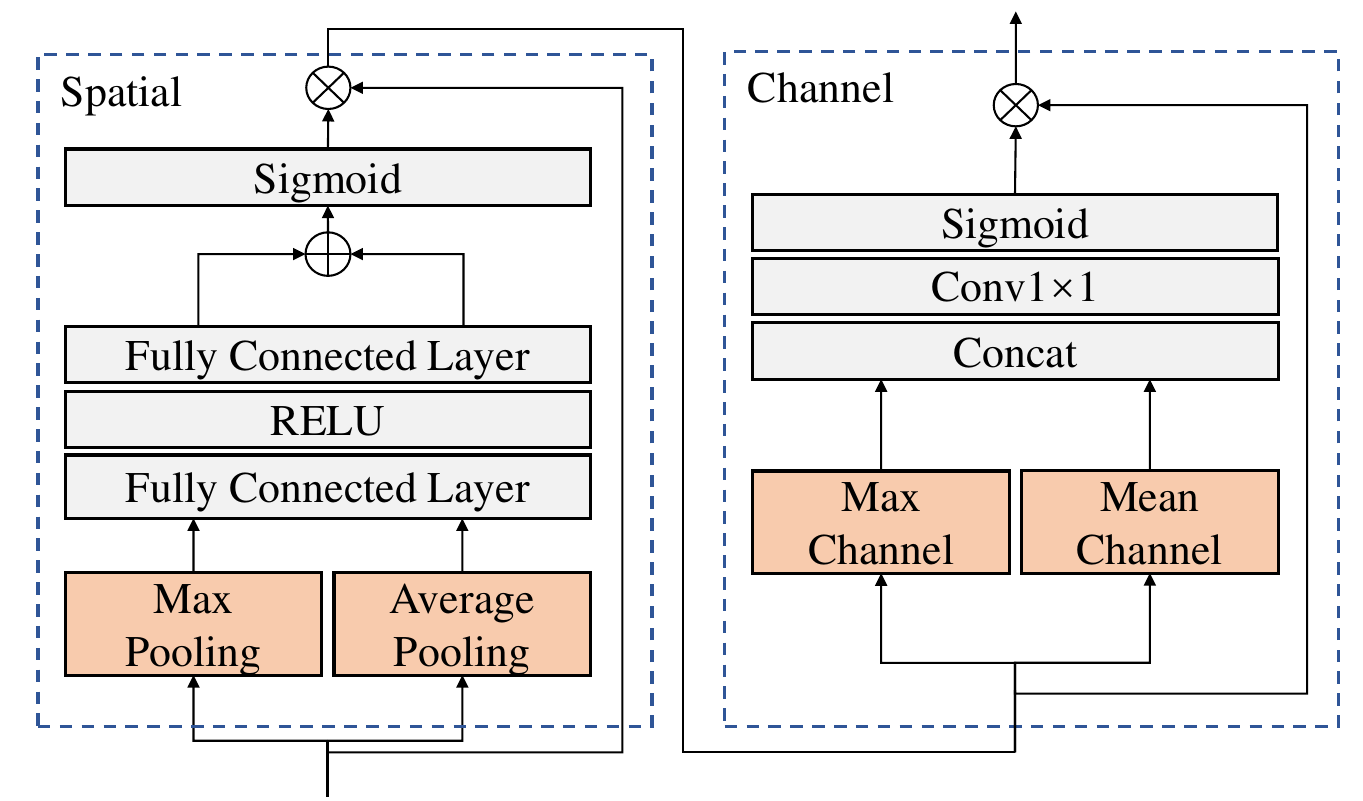}
	\caption{Details of spatial and channel attention structure.}\label{fig.5}
\end{figure}
\begin{align}
x_{s_max}&=FC(RELU\left(FC\left(MaxPooling(x)\right)\right))\\
c_{s_average}&=FC(RELU\left(FC\left(AveragePooling(x)\right)\right)) \\
x_{s_output}&=x*Sigmoid(x_{s_max}+x_{s_average}) \\
x_{c}&=Conv\left(Concat\left(MaxChannel\left(x_{s_{output}}\right),MeanChannel(x_{s_{output}})\right)\right) \\
x_{output}&=x_{s_output}*Sigmoid(x_c)
\end{align}
Where $MaxPooling$ and $AveragePooling$ stand for maximum and average pooling operations respectively, $MaxChannel$ and $MeanChannel$ stand for maximum and average operations on channels, $FC$ stands for fully connected layer, $RELU$ and $Sigmoid$ stand for activation functions, and $*$ stands for multiplication calculation.

\subsection{Decoder}
In the decoder section, the main structure is formed by the UpSample block, wherein the feature maps extracted from the encoder are first adjusted in the number of channels by a simple convolution block. Subsequently, they are spliced with the feature maps from the previous layer on the channels and fused with the features by the convolution block. Additionally, deep supervision is achieved by utilizing 1 × 1 convolution to generate additional segmentation results, enhancing the performance of the model. The structural details are shown in Fig. 6.
\begin{figure}[!t]
	\centering
	\includegraphics[scale=0.5]{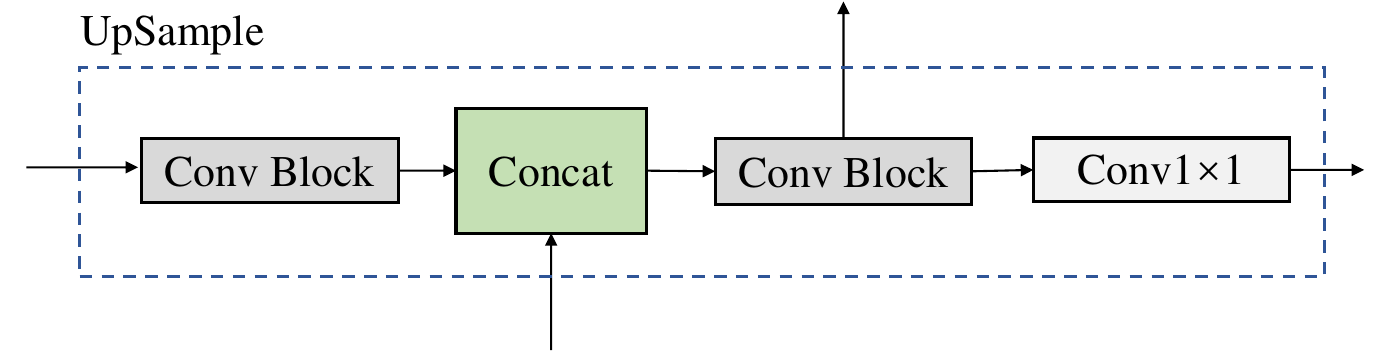}
	\caption{Structure of the decoder block.}\label{fig.6}
\end{figure}

\section{Experiment}
\subsection{Dataset}
The representative and publicly available CrackSeg9K dataset, encompassing several different crack datasets, was chosen as the main experimental dataset. This dataset was re-screened due to several inherent issues. Additionally, experiments were conducted on the private sewer pipe crack dataset, SewerCrack, to demonstrate the model's effectiveness and efficiency in crack segmentation. Lastly, to illustrate the model's generalizability, it was tested on the crack-like retinal blood vessel dataset, CHASE\underline{~}DB1. Specific details of each dataset are described below.

\subsubsection{CrackSeg9K}
Most of the datasets on cracks suffer from a limited number of images, with some containing fewer than 100 images. This scarcity of data can impede both model learning and application. The authors of CrackSeg9k \cite{27} researched 10 crack datasets and constructed a new benchmark crack dataset containing all 10 datasets, which have been listed in Table 1, including the data without cracks. This new dataset includes various crack morphologies such as linear, branching, and reticulated cracks, with each image sized at 400 × 400.
However, upon organizing and deploying the publicly available CrackSeg9k dataset, it was discovered that the dataset published by the authors had quantity and matching issues, likely due to oversight during upload. Consequently, the existing dataset was reorganized and filtered, resulting in 8751 images. Since 5-fold cross-validation is employed for model training, no additional validation sets are required. Therefore, the training set and validation set were combined to form a new training set, adhering to the segmentation ratio specified in CrackSeg9k \cite{27}. This was done to augment the model's learning data. Consequently, the training set to the test set ratio was determined to be 95\% to 5\%. The selection of the test set preceded the selection of the training set, following a specific method. First, the total number of images in the dataset was determined, followed by the calculation of the number of images needed to be selected from this dataset as the test set. The final filtered quantities for each dataset are presented in Table 1. Subsequently, the selection interval was determined by dividing the total number by the number of test sets. This process can be formally expressed as the following formula:

\begin{align}
n_i&=N_i\times5\%\\
m&=\frac{N_i}{n_i}
\end{align}

\begin{table}[!ht]
	\centering
	\caption{Training and testing set partitioning for various datasets.}
	\begin{tabular}{ccc}
		\hline
		Dataset Name & Train & Test \\ \hline
		SDNET2018 & 228 & 12 \\ 
		Ceramic & 95 & 5 \\ 
		CFD & 112 & 6 \\ 
		Crack500 & 2969 & 157 \\ 
		CrackTree200 & 167 & 8 \\ 
		DeepCrack & 421 & 22 \\ 
		GAPS384 & 364 & 19 \\ 
		Masonry & 228 & 12 \\ 
		Rissbilder for Florian & 1893 & 99 \\ 
		Volker & 497 & 26 \\ 
		Noncrack & 1341 & 70 \\ 
		\cline{1-3}
		CrackSeg9K(ours) & 8315 & 436 \\ 
		SewerCrack(ours) & 552 & 30 \\ 
		CHASE\_DB1 & 20 & 8 \\ \hline
	\end{tabular}
\end{table}

\subsubsection{SewerCrack}
The data utilized in this study were derived from a series of raw CCTV videos provided by a municipality in the southern United States, documenting various conditions within sewer pipes. These videos capture a complex environment featuring multiple defects, such as cracks, collapses, and obstructions. The pipes, primarily made of concrete, are circular in shape and range in size from 250 mm to 900 mm in diameter. The original CCTV video format is MPEG-1, with a resolution of 320 × 240 and a frame rate of 30 Frames Per Second(FPS). To extract images from these videos, frames were sampled every 30 frames, resulting in a substantial amount of image data depicting defects in underground water pipes. Frames showing cracks were selected as our original dataset, and manual labeling was performed to create the ground truth values for crack segmentation. This process yielded 582 images with ground truth annotations. Following the segmentation criteria of the reconstructed CrackSeg9k dataset, the data were divided into training and test sets in the same ratio, specifically, 552 images for the training set and 30 images for the test set, as shown in Table 1. Finally, to facilitate model training, the images in the dataset were standardized to a size of 400 × 400 pixels.

\subsubsection{CHASE\underline{~}DB1}
A crack-like segmentation task dataset, CHASE\underline{~}DB1 \cite{28}, was selected to evaluate generalization. This dataset comprises 28 retinal images (999 × 960) from the left and right eyes of 14 school children. Each image was independently and manually labeled by two experts. Following the principle of using the first expert's labeling results, the first 20 images were designated as the training dataset, while the remaining 8 images were used as the test dataset.

\subsection{Implementation details}
Our model was implemented on the well-established nnUNet \cite{29} framework, whose self-configuration feature allowed the focus to remain on designing the network structure rather than other trivial details. The loss function employed is the sum of the Dice loss and cross-entropy loss functions, and deep supervision is performed at each scale. Following the settings in Swin-UMamba, the AdamW optimizer with a weight decay of 0.05 was used. A cosine learning rate decay was applied with an initial learning rate of 0.0001. The pre-training weights from VMambaV2 were used to initialize the encoder part of the model. During the training process, the pre-trained model parameters were frozen for the first 10 epochs to align the other modules. A total of 500 epochs were chosen for training, with the best-performing epoch selected for evaluation. In addition, the evaluation metrics code from CrackSeg9k was used from start to finish for the crack dataset in order to fairly compare the final performance. The configuration of the model is consistent on all datasets, and proper hyperparameter tuning for specific different tasks may yield better results, which is our future work.

\subsection{Compare with other models}
\subsubsection{Crack segmentation}
To demonstrate the superiority of our Mamba-based models, two CNN-based models, UNet \cite{30} and DeepLabV3 \cite{13}, as well as two Transformer-based models, SwinUNet \cite{31} and TransUNet \cite{32}, were selected for comparison. The performance of all models across different datasets was compared while maintaining a consistent input size. As shown in Table 2, our approach achieves the best performance results in both mean Intersection over Union(mIoU) and F1-score. On the reconstructed Crackseg9k dataset, an improvement of 1.6 percentage points on the mIoU metric and 2.16 percentage points on the F1-score metric was observed over the baseline. Furthermore, a comparative implementation was performed on the self-built SewerCrack dataset, which consists of images from the sewer pipe with a complex background and numerous defects, including cracks. Despite the increased segmentation difficulty, our model achieved good performance, improving the mIoU metric by 1.83 percentage points and the F1-score metric by 3.66 percentage points relative to the baseline.

\begin{table}[!ht]
	\centering
	\caption{Results of different models on the reconstructed CrackSeg9k testing dataset and SewerCrack testing dataset. }
	\resizebox{\linewidth}{!}{
	\begin{tabular}{ccccccccc}
		\hline
		\multirow{2}*{Model} & \multirow{2}*{Backbone}  & \multirow{2}*{Input Size} & \multirow{2}*{Params(M)}  & \multirow{2}*{FLOPs(G)} & \multicolumn{2}{c}{CrackSeg9k}  & \multicolumn{2}{c}{SewerCrack} \\
		\cline{6-7}
		\cline{8-9}
		&&&&& mIoU(\%) & F1-score(\%) & mIoU(\%) & F1-score(\%) \\
		
		\hline
		UNet & UNet & 448×448 & 31 & 167 & 76.29 & 70.38 & 58.49 & 31.33 \\ 
		DeepLabV3 & ResNet101 & 448×448 & 59 & 68 & 79.04 & 74.67 & 62.31 & 40.87 \\ 
		SwinUNet & Swin Transformer & 448×448 & 27 & 23 & 78.79 & 74.36 & 61.73 & 39.60 \\ 
		TransUNet & ResNet50+ViT & 448×448 & 93 & 98 & 79.87 & 75.96 & 68.31 & 54.64 \\ 
		Swin-UMamba & VMamba & 448×448 & 55 & 134 & 80.15 & 77.16 & 68.61 & 55.20 \\ 
		CrackMamba & VMamba2 & 448×448 & 75 & 147 & \textbf{81.75} & \textbf{79.32} & \textbf{70.47} & \textbf{58.86} \\ \hline
	\end{tabular}
	}
	\begin{tablenotes}
		\footnotesize
		\item Note: Bolded is the best result.
	\end{tablenotes}
\end{table}

Theoretically, the number of parameters in Mamba-based models is less than in Transformer models(SwinUNet and TransUNet). However, the table indicates that SwinUNet has the smallest number of parameters. This discrepancy arises because SwinUNet employs a tiny model configuration, resulting in a smaller parameter count, which in turn leads to its performance being even lower than that of the CNN-based DeepLabV3. The original Swin-UMamba parameter count is already approximately half that of TransUNet. Our model, adopting a three-branch structure with two Mamba blocks, shows only a modest increase of 20M in the number of parameters, including those for the convolutional branches and other improved structures. All the above theories and results demonstrate that Mamba-based models can maintain a low parameter count while ensuring superior performance.

\subsubsection{Retinal vessel segmentation}

To demonstrate the generalizability of the model, we conducted comparison experiments on a crack-like retinal vessel segmentation dataset. Utilizing the nnUNet architecture, our model automatically preprocesses the images and adjusts them to the most suitable scale size, with an input scale size of 512 × 512. Other models adopt different processing strategies for the original images. For example, FR-UNet \cite{33} employs a sliding window of 48 × 48 to increase the training dataset on the original size image and uses the original size for testing. MERIT-GCASCADE \cite{34} crops the image to 960 × 960 for the input scale, while PVT-GCASCADE \cite{34} uses 768 × 768 and 672 × 672 as input scales. Our model, without using a large input scale or data augmentation, only trains on 20 images and tests on 8 images. Despite this, it achieves performance similar to the other models. Notably, as can be seen in Table 3, our model's IoU metric surpasses MERIT-GCASCADE by 27 percentage points, which is a significant improvement. This substantial increase demonstrates the effectiveness of our model in extracting snake-like features.

\begin{table}[!ht]
	\centering
	\caption{Results of different models on the CHASE\underline{~}DB1 testing dataset.}
	\begin{tabular}{cccc}
		\hline
		Model & F1-score & Sensitivity & IoU \\ \hline
		UNet & 78.98 & 76.50 & 65.26 \\ 
		UNet++ & 80.15 & 83.57 & 66.88 \\ 
		FR-UNet & 79.64 & \textbf{87.98} & 68.82 \\ 
		PVT-GCASCADE & 82.51 & 85.84 & 70.24 \\ 
		MERIT-GCASCADE & \textbf{82.67} & 84.93 & 70.50 \\ 
		CrackMamba & 81.42 & 84.85 & \textbf{97.38} \\ \hline
	\end{tabular}
	\begin{tablenotes}
		\footnotesize
		\item Note: Bolded is the best result.
	\end{tablenotes}
\end{table}

\subsection{Visualization results}

To more intuitively observe the specific performance of the models, a qualitative analysis is conducted by visualizing the segmentation results of the different models on the reconstructed CrackSeg9k dataset. It is evident from Fig. 7 that our model extracts more detailed crack bifurcation structures and removes noise interference in the image. This improvement is attributed to our proposed three-branch fusion module. The snake branch accurately identifies the branching features belonging to the cracks and eliminates noise interference during fusion, preventing the model from focusing on irrelevant noise details. The convolutional branch enhances the model's extraction of local details. The final set of data also demonstrates that the Mamba-based model's strong feature extraction capability for cracks can segment regions that are not marked in the ground truth but are real. Moreover, our improved model aligns even more closely with the actual values. Additionally, it is important to note that both our model and SwinUMamba are based on the nnUNet framework, which stipulates that only pixel values of 1 or 0 can exist in the ground truth during training. All values greater than 0 in the ground truth image are treated as crack pixels, and the model predictions will only include values of 1 or 0. These values are then converted to 255 for visualization purposes. The above reason makes the segmentation results appear visually coarser for our model compared to others, whose predicted values include a range between 0 and 255. However, the same images are used for training and prediction, and the visual difference is merely a result of this conversion process.
\label{fig.7}
\begin{figure}[!t]
	\centering
	\includegraphics[scale=0.3]{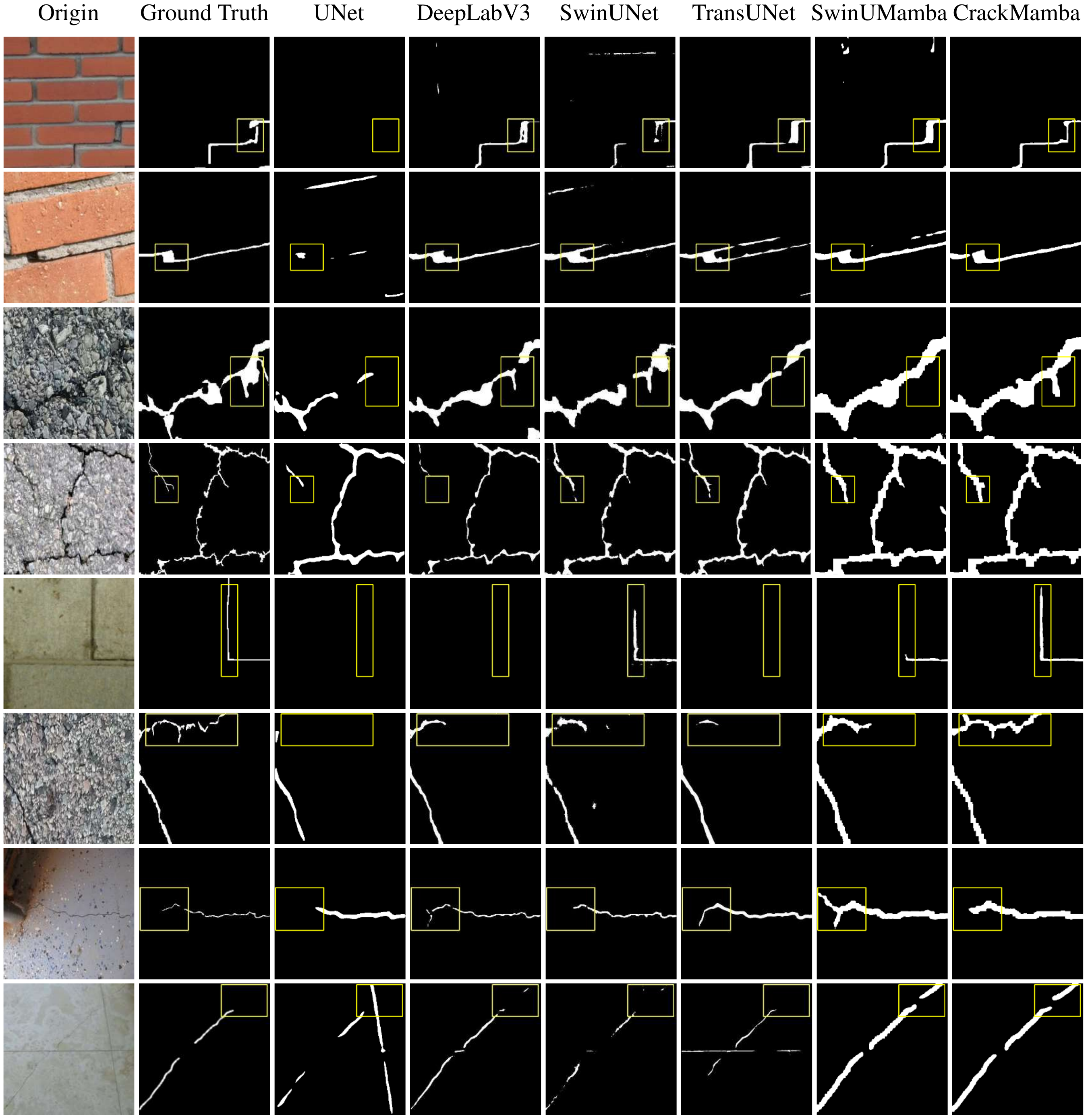}
	\caption{Visualization results of different models on CrackSeg9k.}\label{fig.7}
\end{figure}

\label{fig.8}
\begin{figure}[!t]
	\centering
	\includegraphics[scale=0.3]{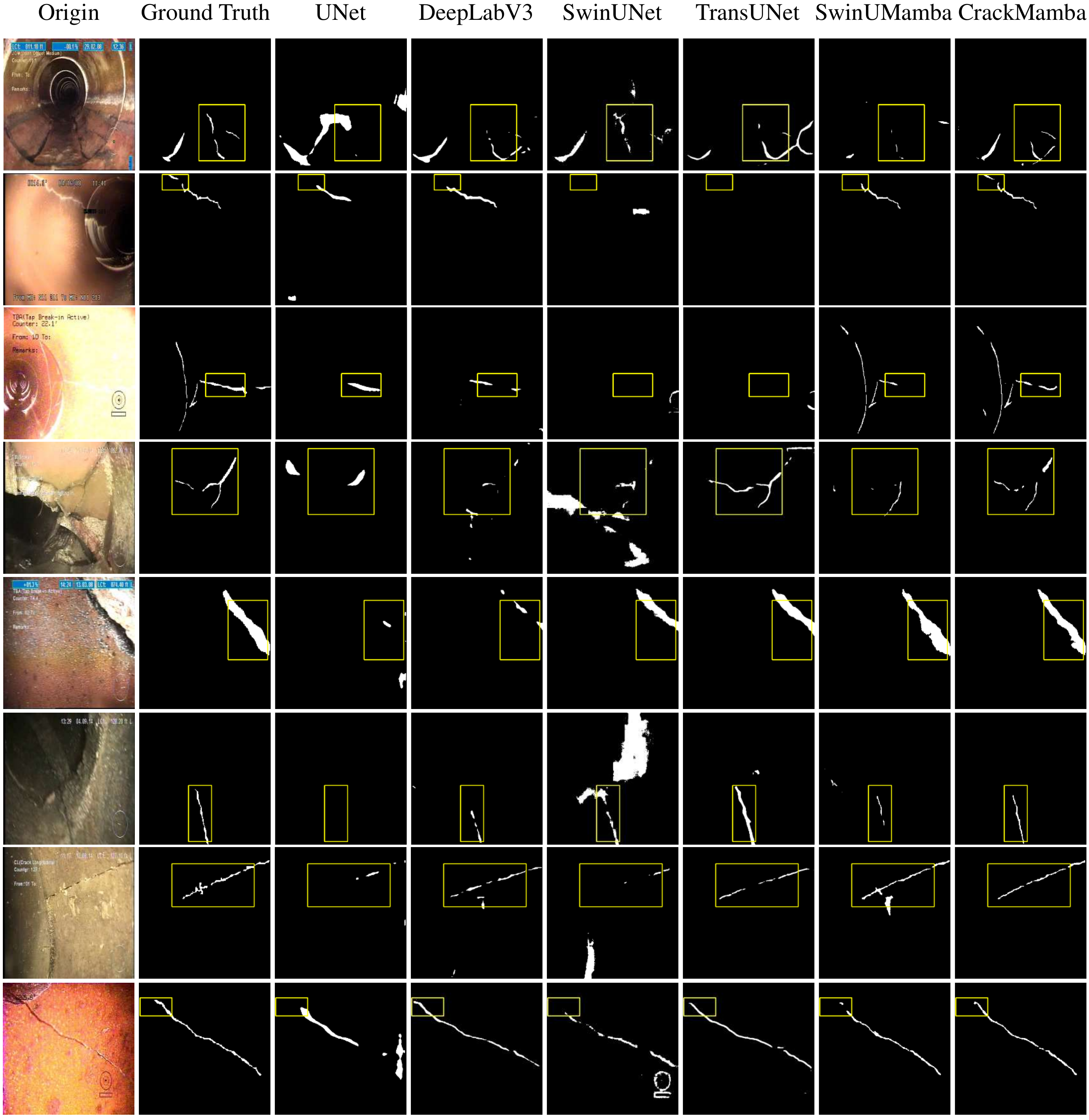}
	\caption{Visualization results of different models on SewerCrack.}\label{fig.8}
\end{figure}
\label{fig.9}
\begin{figure}[!t]
	\centering
	\includegraphics[scale=0.3]{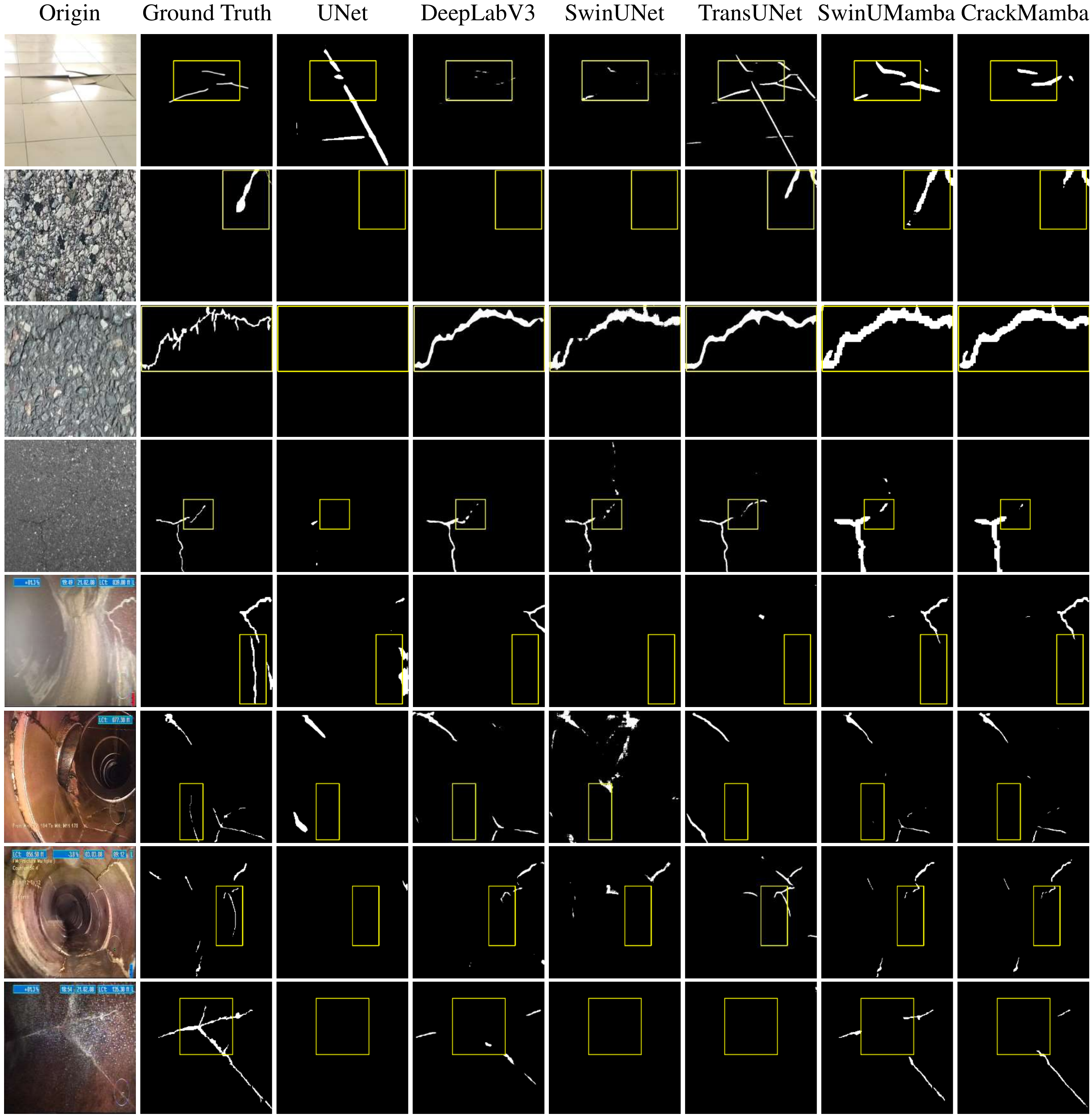}
	\caption{Illustrative examples of suboptimal segmentation performance, comprising four representative plots each from the CrackSeg9K and SewerCrack datasets.}\label{fig.9}
\end{figure}
In addition, the same evaluation was performed on the SewerCrack dataset to visualize the results. As shown in Fig. 8, due to the complexity of the image background, both CNN-based and Transformer-based models encounter difficulties in recognizing the features in the images. For instance, in the second and third rows of data, some crack features are segmented by the CNN-based model, while the Transformer-based model fails to recognize the crack features entirely. Conversely, in the sixth row of data, the CNN-based model struggles to recognize the crack features, whereas the Transformer-based model identifies the crack features but also captures a significant amount of complex background noise. It is worth mentioning that our improved Mamba-based model can recognize the crack features in the images very well, even in the presence of complex background interference. This fully demonstrates the superiority of our model.

Upon visualization of various final segmentation results, instances of suboptimal performance were observed, as illustrated in Fig. 9, where a sequence of four images from both the CrackSeg9K and SewerCrack datasets is presented. Three key conclusions can be drawn from these observations. Firstly, the SCVSS module, which employs Snake scanning and cross scanning techniques, is designed to emphasize feature continuity. While this approach is beneficial for crack detection, it is equally applicable to background feature extraction. Consequently, the model's performance is compromised when minimal contrast between background and crack features is present, or when these features share similarities. This phenomenon is evidenced in the second row of Fig. 9, where cracks are barely discernible to the naked eye. Similar challenges are exhibited in the first and fifth rows, where background elements resembling cracks are observed. The difficulty in identifying fine cracks, as shown in the fourth row, can also be attributed to these factors. Secondly, notable labeling inconsistencies were identified in the CrackSeg9K dataset. For instance, in the third row of Fig. 9, shadows resulting from lighting conditions are incorrectly labeled as crack pixels, leading to reduced IoU scores being recorded by the model. Finally, complex backgrounds are recognized as the most significant challenge in crack segmentation, particularly in sewer environments where multiple defects are often found to coexist. Cracks and collapses are frequently observed in close proximity, further complicating the segmentation task. This complexity is demonstrated in the final rows of Fig. 9, where substantial background noise, an unavoidable consequence of the environmental conditions, is evident in the last row of data. These findings serve to underscore the complexities involved in crack segmentation and highlight areas where potential improvements in both dataset quality and model design can be pursued.

\subsection{Ablation}
\subsubsection{Impact of scanning method}
\begin{table}[!ht]
	\centering
	\caption{Comparative experiments of different scanning methods on the re-screened CrackSeg9K dataset.}
	\begin{tabular}{ccc}
		\hline
		Method & mIoU(\%) & F1-score(\%) \\ \hline
		Cross Scan & 80.92 & 78.19 \\ 
		Random Scan & 81.04 & 78.32 \\ 
		~ & 80.86 & 78.09 \\ 
		~ & 81.21 & 78.56 \\ 
		Snake  Scan & 81.18 & 78.52 \\ \hline
	\end{tabular}
	\begin{tablenotes}
		\footnotesize
		\item Note: All results were obtained using only a single scanning method.
	\end{tablenotes}
\end{table}
We believe that the sequence order of image input to Mamba for feature extraction is crucial and that having target semantic features more adjacent to each other enhances model performance. To verify our theory, a randomized sequence scanning method, Random Scan, was designed. Specifically, the image sequences are scanned randomly, without following a specific order, before being input into Mamba. The processed sequences are then reduced to 2D images. To maximize randomness, two of the sequences in each image are arranged in a random order, with the likelihood of semantic features being adjacent also randomized, while the remaining two sequences are in reverse order. Three experiments were conducted to evaluate the performance of the random scanning method. It is worth noting that, for the sake of convenience, all experiments were conducted using VMambaV2 as the encoder and validated on the re-screened CrackSeg9K dataset. As shown in Table 4, there is significant variability in the performance of the random scanning method. Performance may be lower than the original Cross Scan or higher than our proposed Snake Scan. This variability aligns with our theory, demonstrating that sequence order significantly impacts model feature extraction. Additionally, it underscores the superiority of our Snake Scan method for crack feature extraction. Given the instability of random scanning, Cross Scan and Snake Scan were ultimately selected to form our SCVSS module.

\subsubsection{Impact of improved components}
\begin{table}[!ht]
	\centering
	\caption{Exploring the effect of different improvement modules on model performance on the re-screened CrackSeg9K dataset.}
	\resizebox{\linewidth}{!}{
		\begin{tabular}{cccccccccc}
			\hline
			Groups & V2 & Cross & Snake & Conv & SCA & Params(M) & FLOPs(G) & mIoU(\%) & F1-score(\%) \\ \hline
			1 & ~ & \checkmark & ~ & ~ & ~ & 55 & 134 & 80.74 & 77.93 \\ 
			2 & \checkmark & \checkmark & ~ & ~ & ~ & 65 & 140 & 80.92 & 78.19 \\ 
			3 & \checkmark & ~ & \checkmark & ~ & ~ & 65 & 140 & 81.18 & 78.52 \\ 
			4 & \checkmark & \checkmark & \checkmark & ~ & ~ & 73 & 145 & 81.30 & 78.78 \\ 
			5 & \checkmark & \checkmark & \checkmark & \checkmark & ~ & 75 & 147 & 81.45 & 78.87 \\ 
			6 & \checkmark & \checkmark & \checkmark & \checkmark & \checkmark & 75 & 147 & 81.75 & 79.32 \\ \hline
		\end{tabular}	}
\end{table}
To explore the effectiveness of each improvement, the performance changes were observed by incrementally adding each component to the baseline on the re-screened CrackSeg9k dataset. As shown in Table 5, the selected baseline, referred to as group 1, achieves a mIoU of 80.74\%, which already demonstrates good performance. Replacing the encoder with the better-performing VMambaV2 improves the mIoU by 0.18 percentage points, validating VMambaV2's feature extraction capabilities. Furthermore, the proposed Snake Scan method improves the mIoU by 0.26 percentage points compared to the original Cross Scan. This aligns with the theory that closer semantic neighbors lead to better feature extraction. While the serpentine approach includes two sequences whose order may not match the cracks, resulting in a modest performance improvement, it remains a superior choice compared to the more sparse Cross Scan method. Consequently, fusing Cross Scan and Snake Scan yields significantly better model performance than using either method alone, as demonstrated by the fourth group of data. Additionally, to enhance the model's local feature extraction, a convolutional branch was added to the new fusion module, improving the model's performance by 0.15 percentage points. To better integrate the processing structure of the three-branch approach, the SCA module was included to further process the extracted feature maps, resulting in a final mIoU of 81.75\%.

\section{Limitations and future works}
Our proposed Mamba-based crack segmentation model has demonstrated improvements in both computational efficiency and performance. However, certain limitations persist. As evidenced by our analysis of the visualization results, the model's performance is adversely affected by three primary factors: the similarity between crack and background pixels, errors in ground truth labeling, and complex backgrounds. 

Of these, addressing the inaccuracies in ground truth labeling is our immediate priority. Our next step involves a comprehensive review of the dataset, eliminating incorrect labels and re-annotating as necessary to ensure the model's positive learning trajectory. Secondly, to address the challenge of distinguishing between similar crack and background pixels, we are exploring the integration of deep texture information extraction methods, such as wavelet variations. These techniques could potentially enhance the fusion of crack features, leading to more accurate segmentation results. Finally, to improve segmentation performance in complex backgrounds, we plan to expand our sewer pipe crack dataset and incorporate crack datasets from diverse environments. This approach aims to expose the model to a broader range of scenarios during training, thereby enhancing its generalization capabilities.

In the near future, we are considering the public release of our crack segmentation dataset, which encompasses more complex contexts than the SewerCrack dataset that is contingent upon favorable conditions. This initiative would contribute to the research community significantly and potentially accelerate advancements in the field of crack segmentation.

\section{Conclusion}

In this paper, a Mamba-based crack segmentation is proposed by exploiting two distinct advantages of VMamba: the low computational effort that grows linearly with sequence length and the long-range dependence property of the global receptive field. To further target the application to cracks, the scanning method for converting 2D images into sequences is improved by following the characteristics of the development trajectory of cracks, which is similar to the shape of a snake. A Snake Scan is designed, forming a double-branch structure through a new fusion of the original Cross Scan and Snake Scan, effectively enhancing the model's extraction of global feature information. Additionally, extra convolutional branches were added to give the model the ability to integrate local and global feature information. Experiments on the CrackSeg9k and SewerCrack datasets have fully demonstrated the effectiveness of the proposed model. Additionally, the model's generalization capability has been validated on the retinal vessel segmentation dataset in the medical domain, where it achieved excellent results in tasks similar to crack segmentation. This generalization ability indicates potential applications of the model in various crack segmentation tasks, such as building and road maintenance inspection, as well as quality control of industrial products. Furthermore, the applicability of the Mamba framework in image segmentation has been further confirmed, with the most critical step being the transformation of as many useful features as possible from a 2D image into a compact and proximate 1D sequence. 

The primary contribution of this paper is the development of a crack-specific ordering method, along with the corresponding plug-and-play network blocks designed to fuse global and local feature information to improve crack segmenation efficiency and algorithm generalizability. However, certain limitations remain. Due to the inconsistent positions and orientations of cracks across different images, the method is only able to aggregate crack features and rearrange them in a relatively generic manner, limiting its flexibility and generalizability. Future work will focus on developing image-specific crack feature ordering methods to further enhance the model's flexibility and adaptability.

\section{Data Availability Statement}
Some or all data, models, or codes that support the findings of this study are available from the corresponding author upon reasonable request.

\section{Acknowledgment}
This work was supported in part by NSF of China under Grant No. 61903164 and in part by NSF of Jiangsu Province in China under Grants BK20191427.






\end{document}